\documentclass[doublecol]{epl2}

\usepackage{graphicx} 
\usepackage{hyperref}
\usepackage{xcolor}
\usepackage{amsmath} 
\usepackage{amssymb} 
\usepackage{amsthm}

\usepackage{pdfpages} 

\usepackage{etoolbox}
\newtoggle{long}
\togglefalse{long}

\newtoggle{arxiv}
\toggletrue{arxiv}

\newcommand{\averageswapdistance}{\left< d \right>}

\title{Swap distance minimization shapes the order of subject, object and verb in languages of the world}
\shorttitle{Swap distance minimization in languages of the world}

\author{Jairo Rios-El-Yazidi \inst{1} \& Ramon Ferrer-i-Cancho\footnote{E-mail: rferrericancho@cs.upc.edu (corresponding author).} \inst{1}}
\shortauthor{J. Rios-El-Yazidi \& R. Ferrer-i-Cancho}

\institute{                    
  \inst{1} Quantitative, Mathematical and Computational Linguistics Research Group, Department of Computer Science, Universitat Politècnica de Catalunya, 08034 Barcelona, Catalonia (Spain)
}

\abstract{
Languages of the world vary concerning the order of subject (S), object (O) and verb (V). The most frequent dominant orders are SOV and SVO, and researchers have tailored models to this fact. However, there are still languages whose dominant order does not conform to these expectations or even lack a dominant order. Here we approach word order variation through swap distance, namely the distance between orders in the permutohedron, a graph where the vertices are the six possible orders of S, O and V, and two orders are connected if they differ by a swap of adjacent constituents. 
Then word order variation becomes a problem of assigning probabilities to word orders so as to minimize the average swap distance. 
Although previous work suggests that swap distance minimization may influence synchronic and diachronic variation as well as word order acceptability, here we provide large-scale cross-linguistic evidence across families, macroareas and sources that word order variation within languages is shaped by the principle of swap distance minimization, even when the dominant order is not SOV/SVO and even when a dominant order is lacking.
}

\begin{document}

\maketitle

\newtheorem{property}{Property}
\newtheorem{definition}{Definition}



\section{Introduction}



Approximately 7,000 languages are spoken in the world \cite{Hammarstroem2016a}. Languages vary concerning the dominant order of subject (S), object (O) and verb (V). After controlling for linguistic family, the most frequent dominant order is SOV (as in Turkish and Japanese) and the second most frequent order is SVO (as in English and Mandarin Chinese) \cite{Hammarstroem2016a}. Researchers have tailored models to the frequency distribution of dominant orders \cite{Cysouw2008a,Austin2025a}. For instance, the standard model of typology postulates a subject before verb (SV) preference and subject before object (SO) preference, 
which predicts that the two most frequent orders should be SOV and SVO. However, not all languages are SOV or SVO (e.g., Welsh is VSO and Kaqchikel is VOS) and languages lacking a dominant order (e.g. Nuer or Warlpiri) challenge the models above.
Indeed, $14.3\%$ of languages ($13.5\%$ of families) with a dominant order are not SOV/SVO while $2.4\%$ of languages ($7.1\%$ of families) lack a dominant order \cite{Hammarstroem2016a}. 
The present article is not about why a particular dominant order, either SOV/SVO or another, is selected \cite{wals-81, Cysouw2008a, Goldin-Meadow2008a, Schouwstra2022a, Gell-Mann2011a, Ferrer2013f,Ferrer2024a} or why certain languages may lack a dominant order \cite{Ferrer2013f,Ferrer2024a}. 
This article targets a fundamental question: what constrains word order variation {\em within} languages, even when the dominant order is not SOV/SVO or even when a dominant order is lacking.

\section{The swap distance minimization principle}

\begin{figure*}[t!]
\centering
\includegraphics[width=0.8\linewidth]{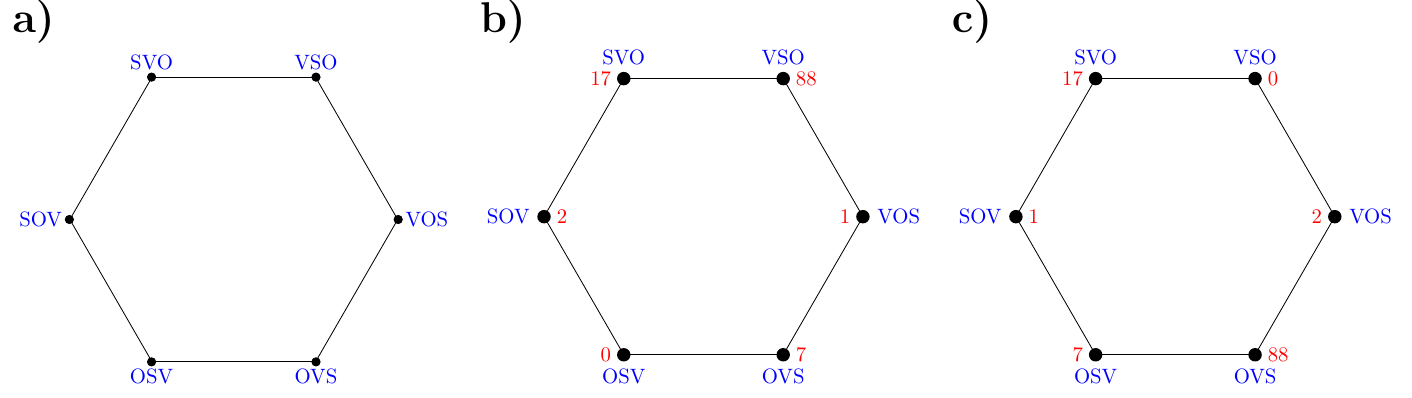}
\caption{a) The permutohedron for the order of S, O and V.
b) The frequency of the order of subject, object and verb in Welsh (Indo-European family) in UD. The average swap distance between pairs of orders is $\averageswapdistance = 0.55$ while the random baseline is $\averageswapdistance_r = 0.7$.
c) A shuffling of the frequencies in Welsh giving 
$\averageswapdistance = 0.88$.
}
\label{fig:permutohedron}
\end{figure*}

There are six possible orders of S, V and O. 
The structure of the space of possible orders can be visualized as a permutohedron, a graph where vertices are orders and an edge indicates that one order can be reached from another by swapping a pair of adjacent constituents (Fig. \ref{fig:permutohedron} a)). 
The swap distance between two orders is defined as the minimum number of swaps of adjacent constituents that are needed to transform one order into another \cite{Franco2024a}.
Then SOV is at distance 0 from itself, at distance 1 from SVO and OSV, at distance 2 from VSO and OVS and at distance 3 from VOS (Fig. \ref{fig:permutohedron} a)). 

It has been hypothesized that the cost of an order variant is a monotonically decreasing function of the swap distance between the variant and the source order \cite{Ferrer2023a}.
The overall cost can be measured by means of 
$\averageswapdistance$, the average swap distance between pairs of orders, that is defined as \cite{Franco2024a}
\begin{equation*}
\averageswapdistance = \sum_{i=1}^{6} \sum_{j=1}^{6} d_{ij} p_i p_j,
\end{equation*}
where $d_{ij}$ is the swap distance between orders $i$ and $j$ and $p_i$ is the relative frequency of the $i$-th order within a language. 

The minimum value of $\averageswapdistance$ is zero and is achieved when a language uses a single order. Its maximum value is $3/2$ and is achieved at least when orders are equally likely. 
Fig. \ref{fig:permutohedron} b) shows the frequency of each order in Welsh (a VSO language).

The principle of swap distance minimization is defined as the minimization of $\averageswapdistance$ \cite{Franco2024a}. The principle does not
dictate which word order is preferred but how word order should vary on top of independent word order preferences.
Preliminary evidence suggests that the principle shapes synchronic and diachronic word order variation \cite{Franco2024a,Ferrer2016c} as well as word order acceptability \cite{Ferrer2023a}.
Here we aim to find large-scale evidence of the action of this principle. Consequently, we need a random baseline, i.e. $\averageswapdistance_r$, that is the average value of $\averageswapdistance$ over all possible shufflings of the six $p_i$'s of a language, namely $6! = 720$ shufflings \cite{Franco2024a}. Fig. \ref{fig:permutohedron} c) shows a shuffling of Fig. \ref{fig:permutohedron} b). 
In this setting \cite{Franco2024a},
\begin{equation*}
\averageswapdistance_r = \frac{9}{5}(1 - S), 
\end{equation*}
where $S$ is the Simpson diversity index that is defined as \cite{Sommerfield2008a}
\begin{equation*}
S= \sum_{i=1}^{6} p_i^2.
\end{equation*}
$\averageswapdistance_r$ removes any swap distance bias but preserves the frequency distribution.

In spite of the linguistic motivation of the problem, swap distance minimization is related to broad themes in complex systems.  
First, $\averageswapdistance$ is a measure of word order diversity as traditional indices such as the Simpson index or the Shannon entropy of the orders, i.e. \cite{Franco2024a}
\begin{equation}
H = -\sum_{i=1}^6 p_i \log p_i.
\end{equation}
Interestingly, $H$ coincides with the Bandt-Pompe permutation entropy for the analysis of time series \cite{Bandt2002a}.
$\averageswapdistance$ shares properties with $H$ (both are minimized when only one order has non-zero probability, and maximized, at least, when all orders are equally likely \cite{Franco2024a}).
However, $\averageswapdistance$ is more powerful: configurations with the same $H$ (or same $S$) may vary concerning $\averageswapdistance$ due to how frequencies are distributed on the permutohedron \cite{Franco2024a}. Such a key difference has implications for information-theoretic approaches to communication, where a principle of entropy minimization for individual words or blocks of words has been postulated \cite{Ferrer2015b,Franco2024a}. Swap distance minimization may be a stronger principle than block entropy minimization \cite{Franco2024a}.
Second, swap distance minimization has implications for how complex systems, e.g., languages and other natural communication systems, are optimized. In particular, the minimization of $\averageswapdistance$ by finding the optimal assignment of the $p_i$'s to vertices of the permutohedron has been shown to be a particular case of the Koopmans-Beckmann Quadratic Assignment Problem (KB-QAP) \cite{Ferrer2025c}. KB-QAP consists of assigning $N$ activities (word order probabilities in our case) to $N$ locations (vertices of the permutohedron in our case) so as to minimize the total cost \cite{Koopmans1957a}. Interestingly, two other optimization problems in communication systems, i.e. compression (the minimization of the average word length with prescribed word frequencies and word lengths) \cite{Petrini2022a} and the minimization of the distance between syntactically related words \cite{Ferrer2020b}, turn out to be particular cases of KB-QAP and thus a general optimal assignment principle has been postulated \cite{Ferrer2025c}.

Here we will show that languages tend to satisfy $\averageswapdistance < \averageswapdistance_r$ within linguistic families from all over the planet more often than expected by chance.  

\section{Materials}

\subsection{The sources for word order frequencies}

We obtain the frequency of the order of subject, object and verb from two main sources: the New Testament (NT) \cite{Oestling2015a} and syntactic dependency treebanks from the Universal Dependencies (UD) collection \cite{Sanguinetti2026a_Encyclopedia}.
The latter are used to ensure that results are not a consequence of the automatic methods used to retrieve subjects, objects and verbs in NT.
These treebanks have the potential to capture subject, object and verb more accurately than NT but covering fewer languages and thus increasing the risk of undersampling.

As for NT, we employ ready-to-use counts from the New Testament (NT) \cite{Oestling2015a}, available from \url{https://aclanthology.org/attachments/P15-2034.Datasets.zip}.
The dataset provides token-based counts and type-based counts \cite{Oestling2015a}. Here we use only token-based counts.
As for UD, we also use token-based counts derived from the latest version of the Universal Dependencies (UD) collection of treebanks \cite{Nivre2020a}, namely UD 2.17, that is available from \url{https://universaldependencies.org/download.html}.

To control for syntactic annotation, we combine two competing syntactic annotation criteria, i.e. UD and Surface-Syntactic Universal Dependencies (SUD) \cite{Sanguinetti2026a_Encyclopedia}.
The UD 2.17 collection provides its own annotation style, that is a content-head annotation style \cite{Sanguinetti2026a_Encyclopedia}. 
We also consider the corresponding treebanks following a function-head annotation style: the SUD 2.17 treebank collection \cite{sud}. The latter treebanks are available from \url{https://surfacesyntacticud.org/data/}.

All treebanks of the same language are merged into a single treebank. We include all treebanks in the SUD collection, including the so-called native SUD treebanks. Then the SUD collection has initially more treebanks than the UD collection. After applying the preprocessing steps described below, we obtain a sample of 953 languages from 111 linguistic families for NT and a sample of 170 languages from 31 linguistic families both for UD and SUD.

\subsection{Retrieval of linguistic family and macroarea of a language}

We use Glottolog to determine the linguistic family of languages and their macroarea \cite{Hammarstroem2022a}, in particular Glottolog 5.3 that is available from 
\url{https://doi.org/10.5281/zenodo.3260727}.
Languages whose family is indeed a pseudofamily (sign language, pidgin, artificial language) were excluded from the analysis. 
If a language was identified as an isolate or it was itself the root of a family, its own name became the family name. Thus each isolate becomes the only member of a family. 


\subsection{The typical macroarea of a family}

To measure the degree of association of a family with a macroarea, we calculate a weighted proportion of languages of the family within each macroarea. 
For a family and a macroarea $\mu$ this proportion is formally defined as
\begin{equation}
\tau(\mu) = \frac{1}{T}\sum_{i \in F_\mu} \frac{1}{a_i},
\end{equation}
where $T$ represents the total number of languages of the family in the sample, $F_\mu$ is the set of languages of the family in the macroarea $\mu$ and $a_i$ is the number of macroareas covered by the $i$-th language of that family. The typical macroarea of a family is the macroarea that maximizes $\tau(\mu)$.

\subsection{Detection of S, O and V triplets in treebanks}

Triplets formed by S, O, and V were extracted from treebanks searching for verbs that have at least two dependents: a nominal subject and a nominal (direct) object (as opposed to clausal subject or clausal object). 
We focus on nominal subjects and nominal objects for several reasons. 
First, for consistency with the definition of dominant order \cite{wals-s6} and research on dominant word order \cite{Choi2021a, Ehara2025a}. Second, to reduce the chance of competition between swap distance minimization and syntactic factors that determine word order \cite{wals-81} or dependency distance minimization, which is more likely to surpass other principles when the subject or the object are clausal \cite{Ferrer2019a,Ferrer2023b}. Third, to align with the low syntactic complexity of the stimuli of acceptability experiments where preliminary evidence of swap distance minimization has already been found \cite{Ferrer2023a}.
See the Supplementary online information for technical details on the detection of these triplets in treebanks depending on the annotation criteria.

\subsection{Problem solving and data exclusion}

Certain languages were lost as a result of our preprocessing methods (no subject-object-verb triplet met the selection criteria). Sometimes we encountered data labeled with the name of a dialect or a macrolanguage. 
In those cases, we had to find the right language and decide whether to keep the data or remove it based on the principle that there should be only one sample per language, i.e. each language contributes with just one value of $\averageswapdistance$. 
The justification is that all languages must have the same weight within a family; otherwise, the most represented languages would bias the results of the statistical test that is run within each family. Notice that there are two solutions for obtaining a single point per language: removing the redundant samples or merging the samples. We decide to remove the redundant samples. When removing redundant samples, we prioritize the removal of dialects or macrolanguages over samples labeled as languages. We choose these criteria for simplicity and to avoid risks. 

During this process we had to deal with  inconsistencies across resources, missing or ambiguous identifiers, and mismatches between language names and existing classifications. 
A detailed list of problems and their solution is provided in the Supplementary online information.

Code and data are at an online repository, \url{https://osf.io/wq9pk/overview?view_only=b550555762124eea9ad8009e4e32bff7}.

\section{Methods}

\subsection{The meaning of swap distance}

Swap distance is just a means to measure cost with respect to the permutohedron. The swap distance minimization principle assumes that the cost of producing a variant from a source grows with the minimum number of swaps of adjacent elements that are required to transform the source order into that variant. 
Swap distance is agnostic about the actual mechanism producing the variant during evolution \cite[Supplementary material]{Ferrer2016c} or within a language.  
That is, OVS may be produced in two steps from SOV, swapping a pair of adjacent constituents each time, or it may be produced in just one step by postposing the subject. Indeed, the transition from dominant SOV to dominant OVS in evolution is known to have taken place in one step \cite{Gell-Mann2011a}. Swap distance simply predicts that such a transition is less likely than transitions involving orders at distance 1 (all other things being equal). Indeed, this is what the evolution of word order suggests \cite{Gell-Mann2011a,Ferrer2016c}.

\subsection{The main statistical test}

The core statistical test for swap distance minimization is a one-tailed Wilcoxon signed-rank test, a non-parametric matched-pairs test \cite{Conover1999a} that compares $\averageswapdistance$ against $\averageswapdistance_r$ for each language within a set. The test is borrowed from \cite{Franco2024a}. The test is one-tailed because swap distance minimization predicts $\averageswapdistance < \averageswapdistance_r$. By default, the set is formed by languages from the same linguistic family. We also consider a set formed by languages lacking a dominant order.
A further justification of the choice of this test is provided in the Supplementary online information.

\subsection{Testing on each family}

Given a source (as defined above) the statistical method is the following:
\begin{enumerate}
\item
For each language in the source, we calculate $\averageswapdistance$ and $\averageswapdistance_r$.
\item 
For each family in the source, we run the Wilcoxon signed-rank test to compare $\averageswapdistance$ against $\averageswapdistance_r$ for each language in a family and obtain a $p$-value. 
The $p$-value may be small simply because the trend has been transmitted by a common ancestor. For instance, Romance languages may have received the trend from Latin. But Latin was SOV and Romance languages are SVO hence a key point is the presence of swap distance minimization effects in a family even when the dominant order may have been changing over time or not all languages in the family have the same dominant order. 
\item 
On top of the tests, we adjust the $p$-values obtained for each family to control for multiple comparisons and to check if the trend goes beyond individual families. In particular, we use the step-down minP (sd.minP) correction method \cite{Westfall1993a,Westfall2008a}. See the Supplementary online information for further details on $p$-value adjustment.
\iftoggle{long}{\item 
On top of the adjusted $p$-values, we calculate the number of families with an adjusted $p$-value of $\alpha$ or smaller to inform about the number of families where the swap distance minimization effect is unlikely to be due to chance. We also calculate the number of languages in the families such that their adjusted $p$-value is $\alpha$ to inform about the presence of undersampling within each family (i.e the number of languages representing the family is too small for the statistical test to produce a small $p$-value). }{}
\end{enumerate}
This approach has the advantage of allowing one to count the number of families where swap distance minimization effects are less likely to be due to chance. However, it has several limitations. First, it may underestimate the true count because certain families are represented by a number of languages that is too small for giving a small $p$-value. Second, families with more languages may easily get smaller $p$-values. Critically, $\averageswapdistance < \averageswapdistance_r$ may have been passed to languages in the family by some common ancestor resulting in misleadingly small $p$-values.

The next approach fixes the problem of vertical transmission by sacrificing the ability to identify specific families or to provide family counts.

\subsection{Stratified sampling}

Given a source or a subset of a source (the languages of a source lacking a dominant order), we aim to test for swap distance minimization in a way that cannot be distorted by the size of a family or vertical transmission within the family. Accordingly, 
we estimate the $p$-value of the non-parametric test by means of stratified random sampling with the families of a source as strata \cite{Rijkhoff1998a}. In particular, we estimate a confidence interval (CI) for the $p$-value at $99\%$ confidence with the following procedure. We generate $N_s$ 
samples of languages where there is exactly one language of each family in the sample and the language representing each family is chosen uniformly at random among the languages of the family. This corresponds to variety sampling, where linguistic diversity is maximized \cite{Rijkhoff1998a}.
The CI is computed on the $p$-values of the Wilcoxon signed-rank test obtained for each sample. Here we use $N_s = 10^6$.

\subsection{The dominant order of subject, object and verb of a language}

To assign an order of subject, object and verb to a language, we follow the traditional approach in typology as in the World Atlas of Linguistic Structures (WALS) and related projects: each language is classified into one of the six possible orders of S, O and V or into the special category ``No dominant order" (NDO) to indicate the lack of a dominant order \cite{wals-81,Hammarstroem2016a}. 
We can obtain such information in two ways: querying the WALS database, available from \url{https://doi.org/10.5281/zenodo.13950591}, or applying a statistical method \cite{wals-s6,  Choi2021a, Ehara2025a}. 
If we query the WALS database the problem is that $60.76\%$ of languages in NT do not have information about the dominant order. 
As WALS does not have information about the dominant order for all languages in our datasets, we must use a statistical criterion to obtain that information for all of them.

The traditional statistical criterion is as follows \cite{Choi2021a}. Consider the ratio $\rho = p_2/p_1$, 
where $p_i$ is the frequency of the $i$-th most frequent order. Consider also a threshold $\rho_0$. By definition $0 \leq \rho \leq 1$. 
If $\rho \geq \rho_0$ then the language is classified as lacking a dominant order (NDO); if $\rho < \rho_0$ then the most frequent order is the dominant order. \cite{wals-s6, Choi2021a} chose $\rho_0 = 1/2$ while \cite{Ehara2025a} investigated the effect of $\rho_0$ and found that $\rho_0 \geq 0.5$ warrants an accuracy of $80\%$ or greater in guessing the information about the dominant order of S, O and V in the WALS database. We choose $\rho_0 = 0.5$ for consistency with \cite{wals-s6,  Choi2021a}. If the analysis is restricted to languages having a dominant order, it turns out that the most frequent order and the dominant order agree for $85.7\%$ of the languages in the New Testament dataset \cite[Table 2]{Oestling2015a}. Therefore, the statistical approach gives results that are consistent with the WALS database. 

Although the notion of dominant belongs to the so-called type-based typology, our adoption of a flexible statistical criterion to determine the dominant order is in line with token-based typology \cite{Levshina2019a}. Although these two approaches to typology may look opposite, recent mathematical results on swap distance minimization strongly suggest that the most likely order is indeed the source order from which all other orders emanate, reconciling these two approaches when swap distance minimization is the only principle of word order  \cite{Ferrer2025c}.
Notice that the statistical criterion has no impact on the statistical tests within families because the dominant order is only used for visualization. The statistical criterion has an impact on the statistical power of the analysis within languages lacking a dominant order. 
If we query WALS, the number of NDO languages obtained is 36 in NT, 9 for UD and 9 for SUD. If we use a statistical criterion with $\rho_0 = 0.5$, the number of NDO languages is 268 in NT and 37 in both UD and SUD. The gain in languages of the statistical criterion results in much lower $p$-values. 
We may confirm swap distance minimization in NDO languages trivially because the resulting sample of NDO languages with $\rho_0 = 0.5$ is so broad that it simply reflects the statistical properties of a whole sample formed by NDO languages and languages with a dominant order. Therefore, we consider more restrictive thresholds within the range where the accuracy  in predicting WALS dominant order is maximized ($\sim 85\%$) according to \cite[Figure 1]{Ehara2025a}, i.e. $\rho_0 \in \{0.6, 0.7\}$. 

\section{Results}

\begin{figure*}[t]
\centering
\includegraphics[width = \linewidth]{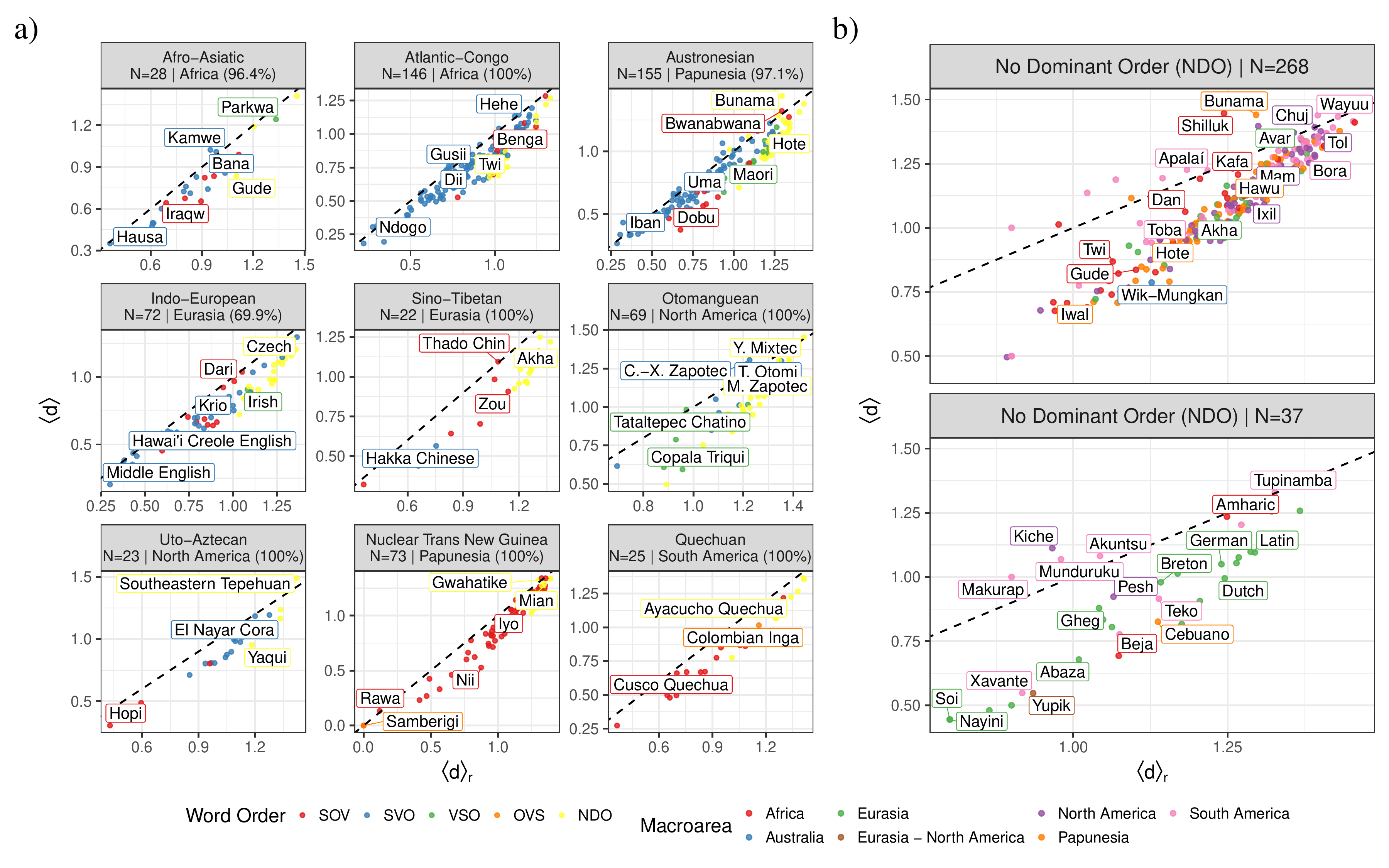}
\caption{\label{fig:NT_combo}
Average swap distance ($\averageswapdistance$) as a function of the random baseline ($\averageswapdistance_r$). Points stand for languages. The dashed line is a control line to indicate $\averageswapdistance = \averageswapdistance_r$. Points below the control line are languages such that $\averageswapdistance < \averageswapdistance_r$.
$N$ indicates the number of languages. a) The linguistic families with lowest $p$-value in NT. Percentages below the family name indicate the percentage of languages in the typical macroarea of the family. 
Languages are colored by their dominant order. In the legend, orders are sorted following the permutohedron (Fig. \ref{fig:permutohedron}).
The name of certain languages of the Otomanguean family had to be abbreviated: Yosondúa Mixtec (Y. Mixtec), Cuixtla-Xitla Zapotec (C.-X. Zapotec), Mitla Zapotec (M. Zapotec) and Tenango Otomi (T. Otomi).
b) Languages lacking a dominant order in NT (top) and UD (bottom). Languages are colored by macroarea. A hybrid macroarea (Eurasia - North America) was created to accommodate Yupik (Eskimo-Aleut), that is spoken in two macroareas.
\iftoggle{long}{({\em C}) The effect of the significance level $\alpha$ in NT (left), UD (center) and SUD (right). Vertical red dashed lines indicate $p$-value milestones. On top, the backward cumulative distribution, namely the number of families with an adjusted $p$-value of $\alpha$ or smaller. Horizontal blue dotted lines indicate the number of families reaching a certain $p$-value milestone (their intercepts are inferred by linear interpolation from the two nearest points). 
At the bottom, the number of languages in the families such that their adjusted $p$-value is $\alpha$. Points stand for distinct families but different families with the same values may be superimposed. Horizontal blue dotted lines indicate the number of languages reached at each $p$-value milestone (their intercepts are inferred by linear interpolation from the two nearest points). Note: after adjusting the $p$-values, certain $p$-values become practically zero due to the Monte Carlo procedure (SI Appendix) and this is in conflict with the logarithmic scale used. For that reason, the values of $\alpha$ for these points are fictitious low numbers. }{}
}
\end{figure*}

We find a trend for $\averageswapdistance < \averageswapdistance_r$ across linguistic families and macroareas in NT (Fig. \ref{fig:NT_combo} a)). 
To assess if the trend is significant within each family, we run a non-parametric test that compares the values of $\averageswapdistance$ and $\averageswapdistance_r$ of the languages of the family. 
Within a family, the trend may be simply due to transmission from a common ancestor. Therefore, we look for further evidence across families.
Nine families in NT have an adjusted $p$-value smaller than $10^{-5}$ (Fig. \ref{fig:NT_combo} a)). Therefore, a low significance level suffices to cover all macroareas except Australia in NT. Notice that there are only five languages from Australia in NT, that are split into the following families (languages in parenthesis): Maningrida (Burarra), Pama-Nyungan (Djambarrpuyngu, Kuku-Yalanji and Wik-Mungkan)
and Indo-European (Kriol). The statistical test fails to give a low $p$-value for the Australian language families due to the low number of languages representing each family. The problem is even more dramatic in UD and SUD, where there is only one Australian language (Warlpiri from Pama-Nyungan).

\iftoggle{long}{
Interestingly, the number of families with a $p$-value not exceeding a certain significance level and the number of languages per family with that level evolve in the opposite direction in all sources (Fig. \ref{fig:NT_combo} c)).
In particular, as the significance level $\alpha$ increases, the number of linguistic families whose adjusted $p$-value is $\alpha$ or lower increases, but the number of languages representing the family decreases, suggesting that undersampling within families causes a loss of statistical power that shadows the effect of swap distance minimization. 
}
As families with more languages can achieve lower $p$-values and inheritance within a family may reduce the $p$-value further, we estimate the $p$-value of the non-parametric test for each source with stratified random sampling (with families as strata) for each source obtaining a $99\%$ confidence interval (CI): $[6.2\times10^{-20},6.1\times10^{-17}]$ in NT, $[2.4\times10^{-5},4.9\times10^{-3}]$ in UD and $[2.2\times10^{-5},2.6\times10^{-3}]$ in SUD. Therefore, the effect of swap distance minimization remains even after controlling for phylogenetic relatedness.
Finally, if we restrict the analysis to languages lacking a dominant order with $\rho_0 = 0.5$, 
we still recover the same trend in NT (Fig. \ref{fig:NT_combo} b) top) as well as in UD (Fig. \ref{fig:NT_combo} b) bottom) and also SUD according to the confidence interval for the $p$-values in Table \ref{tab:NDO}.
In NT, all the six macroareas are covered, while in UD, only Australia is missing (Fig. \ref{fig:NT_combo} b)). Similar conclusions are reached with $\rho_0 \in \{0.6, 0.7\}$ in spite of the reduction in the number of languages classified as NDO (Table \ref{tab:NDO}). 

\begin{table}
\caption{\label{tab:NDO} Swap distance minimization in NDO languages as a function of $\rho_0$, the threshold for classifying a language as NDO. For each threshold and source, we show the number of NDO languages (langs.), the number of families covered and the confidence interval (CI) of the $p$-value of the non-parametric test for $\averageswapdistance < \averageswapdistance_r$ estimated with stratified sampling with families as strata.}
\begin{center}
\begin{tabular}{ccrrr}
$\rho_0$ & source & langs. & families & $99\%$ CI \\ 
\hline
0.5 & NT & 268 & 56 & $[9.7\times10^{-11},2.3\times10^{-8}]$\\
    & UD &  37 & 12 & $[7.3\times10^{-4},8.1\times10^{-3}]$ \\
    & SUD & 37 & 12 & $[4.9\times10^{-4},6.1\times10^{-3}]$\\
0.6 & NT & 170 & 41 & $[7.7\times10^{-11},1.6\times10^{-6}]$\\
    & UD &  24 & 7 & $[7.8\times10^{-3},3.9\times10^{-2}]$\\
    & SUD & 25 & 8 & $[3.9\times10^{-3},2.0\times10^{-2}]$\\    
0.7 & NT & 129 & 36 & $[2.0\times10^{-9},1.9\times10^{-5}]$\\
    & UD &  17 & 7 & $[7.8\times10^{-3},2.3\times10^{-2}]$\\
    & SUD & 21 & 7 & $[7.8\times10^{-3},3.9\times10^{-2}]$\\    
\end{tabular}
\end{center}
\end{table}

\section{Discussion}

We have shown that swap distance minimization shapes word order variation within languages independently of the source and the annotation criterion.
Although the New Testament data offers extensive coverage, it may introduce translation-related or genre-specific effects. The inclusion of UD and SUD data mitigates these potential problems by increasing the accuracy and genre diversity but sacrificing coverage. The key point is not the  limitations of each source but why we eventually confirm the presence of swap distance minimization effects in spite of the limitations of each source.

We have also shown that swap distance minimization shapes word order variation even when the languages are not SOV/SVO as expected by typology models \cite{Cysouw2008a,Austin2025a}. 
This is clearly visible in families with high dominant order diversity but such that dominant SOV is scarce (Fig. \ref{fig:NT_combo} a)), as in Austronesian (SOV in 18 out of 155 languages) and Otomanguean (no SOV),
and crucially, within languages lacking a dominant order (Fig. \ref{fig:NT_combo} b)) even when using more stringent definitions of NDO (values of $\rho_0$ higher than the traditional $\rho_0 = 0.5$ \cite{wals-s6, Choi2021a}).
The key point is not whether the notion of dominant order is appropriate in a statistical or conceptual sense \cite{Levshina2023a} but the fact that swap distance minimization underlies the traditional notion of dominant order. 

Although here our goal is not to explain why a particular order is selected, swap distance minimization may compete with (a) pragmatic or syntactic factors that determine word order in flexible order languages \cite{wals-81}
or (b) other word order principles such as dependency distance minimization or predictability maximization \cite{Liu2017a,Ferrer2013f} when they select orders that are far away in the permutohedron. 
Then it should not be surprising, that across families and macroareas, certain languages exhibit values of $\averageswapdistance$ that are very close to $\averageswapdistance_r$ or even greater (Fig. \ref{fig:NT_combo}). Conflicts between word order principles have already been predicted and confirmed \cite{Ferrer2023b,Ferrer2024a}. The effects of swap distance minimization are widespread across dominant orders, linguistic families and distant areas of the world.  
We have excluded that such a massive phenomenon results only from inheritance within families and demonstrated its presence in families from all macroareas except Australia, due to the severe scarcity of Australian languages in our dataset.

While our stratified sampling by family is an important control for phylogenetic relatedness, we have not fully addressed possible areal dependencies, which is a remaining limitation of our analyses. Thus, a challenge for future research is whether the phenomenon is merely an artifact of language contact, and if so, how just appearance of swap distance minimization would originate in the first place and be preserved during transmission in the absence of the principle itself. 

We have demonstrated that $\averageswapdistance$, an alternative to the
Bandt-Pompe permutation entropy \cite{Bandt2002a}, is able to capture hidden constraints in languages. Therefore, we have shown the potential of indices that are aware of the structure of the permutohedron for future research on time series based on Ordinal Analysis \cite{Leyva2022a}. 

\acknowledgments
We are grateful to C. Bentz for making us aware of \cite{Oestling2015a} in 2015. We thank D. Dediu, A. Hernández-Fernández, J. Troendle and C. Ferrer-i-Menacho for helpful discussions. This research is supported by a recognition 2021SGR-Cat (01266 LQMC) from AGAUR (Generalitat de Catalunya) and the grant PID2024-155946NB-I00 funded by Ministerio de Ciencia, Innovación y Universidades (MICIU), Agencia Estatal de Investigación (AEI/10.13039/501100011033) and the European Social Fund Plus (ESF+).

\bibliographystyle{eplbib}
\bibliography{biblio}

\iftoggle{arxiv}{


\section*{Supplementary material}

# Supplementary online information of "Swap distance minimization shapes the order of subject, object and verb in languages of the world"

Jairo Rios-El-Yazidi & Ramon Ferrer-i-Cancho

## 1 Materials

### 1.1 Problem solving and data exclusion

Below we explain the criteria and solutions adopted to ensure consistency and comparability across datasets. Throughout this subsection, linguistic families are indicated in square brackets to make family assignments explicit and uniform.

#### 1.1.1 Assigning linguistic families to languages in NT

Since this dataset is indexed by ISO 639-3 codes, each language is explicitly identified by its ISO code, which is indicated below in parentheses to avoid ambiguity.

- **Languages where all frequencies are 0, removed.** Some languages in Östling's dataset [1] already have a frequency of zero for all word orders. Therefore, the following samples where excluded from our analyses:
    - Hixkaryána (`hix`) [Cariban]
    - Weenhayek (`mtp`) [Mataguayan]

- **Languages with non-existent ISO codes, removed.** Some entries in the dataset of Östling [1] have invalid or useless ISO codes:
    - `erv`, non-existing code.
    - `qej`, leads to "Reserved for local use".

    After revising and comparing the texts against more recent versions of the corpus with fixed metadata, the new proper codes for each turn out to be [2]:
    - `erv`, Kannada (`kan`) [Dravidian]
    - `qej`, Huamalíes-Dos de Mayo Huánuco Quechua (`qvh`) [Quechuan]

In addition, there are already samples of each two languages in the dataset. The `erv` sample turns out to be Kannada with Kannada script and the `kan` sample corresponds to latinized Kannada. The `qej` sample turns out to be a near duplicate of the `qvh` sample [2]. For these reasons, we decide to remove the `erv` and `qej` to ensure that every language is represented once, prioritizing samples were the initial ISO codes were correct or informative.

- **Macrolanguages without Glottolog entries, removed or classified under the families of their constituent languages.** The ISO 639-3 codes of certain entries in the dataset of Östling [1] correspond to macrolanguages that do not have an entry in Glottolog. If none of the languages they encompass appears separately in the NT collection, they are retained and classified under the highest-level language family:

    - Azerbaijani (`aze`) [Turkic]
    - Estonian (`est`) [Uralic]

  If any of their languages appears separately, they are removed. The removed macrolanguage samples are the following:

    - Malagasy (`mlg`) [Austronesian]. Plateau Malagasy (`plt`) already appears.

  Some entries in the dataset of Östling [1] have ISO codes that are 639-5 (for macrolanguages) instead of the ISO 639-3 (for languages). The samples with the following codes

    - `grk`. ISO 639-5 for "Greek languages".
    - `nah`. ISO 639-5 for "Nahuatl languages".

  were removed since the dataset already contains samples of languages representing these macrolanguages.

- **Macrolanguages listed in Glottolog as low-level families, removed or classified under the highest-level family.** Some ISO 639-3 macrolanguages have entries in Glottolog but are classified as low-level families rather than as individual languages. If none of the languages they cover appears separately in the NT collection, they are classified under the highest-level language family. This is the case of

    - Uzbek (`uzb`) [Turkic]

  If any of their languages appears separately, they are removed. The removed low-level family samples are the following:

    - Buriat (`bua`) [Mongolic-Khitan]. Russia Buriat (`bxr`) already appears.

- **Languages listed in Glottolog as split into two, removed.** These ISO 639-3 codes correspond to languages that have been split into two distinct languages in Glottolog due to significant differences within the original language. Certain samples had to be removed because at least one of the resulting specific languages is already present in the dataset and each language must be represented by a single sample. The samples removed and the reasons of the removal are the following:
    - Falam Chin (Retired) (`flm`) [Sino-Tibetan]. Falam Chin (`cfm`) already appears.
    - Izere (Retired) (`fiz`) [Atlantic-Congo]. Izere (`izr`) already appears.
    - Patla-Chicontla Totonac (`tot`) [Totonacan]. Both Tecpatlán Totonac (`tcw`) and Upper Necaxa Totonac (`tku`) already appear.

- **Languages identified as duplicate entries in Glottolog based on geographic names, removed.** These entries of the dataset are indexed in Glottolog under the names of specific localities rather than the linguistic variety itself. The following samples are removed because the standard linguistic entry for that variety is already present in the collection:
    - Kosena (`kze`) [Nuclear Trans New Guinea]. Awiyaana (`auy`) already appears.

- **Languages listed in Glottolog as dialects, removed or classified under the family of their language.** The ISO 639-3 codes for these languages appear in Glottolog, but the corresponding entries are treated as dialects of larger language units. If their language does not appear separately in the NT collection, they are classified under the family of that language. These dialects and their solutions are the following ($\rightarrow$ indicates the language to which each dialect belongs):
    - Croatian Standard (`hrv`) $\rightarrow$ Serbian-Croatian-Bosnian [Indo-European]
    - Guajajára (`gub`) $\rightarrow$ Tenetehara [Tupian]
    - Serbian Standard (`srp`) $\rightarrow$ Serbian-Croatian-Bosnian [Indo-European]
    - Twi (`twi`) $\rightarrow$ Akan [Atlantic-Congo]

    If their language appears separately, they are removed. The removed dialect samples and the reasons of the removal are the following:
    - Achi', Cubulco (`acc`) [Mayan]. Achi (`acr`) already appears.
    - Chuj, Ixtatán (`cnm`) [Mayan]. Chuj (`cac`) already appears.
    - Huastec, San Luís Potosí (`hva`) [Mayan]. Huastec (`hus`) already appears.
    - Ixil Nebaj (`ixi`) [Mayan]. Ixil (`ixl`) already appears.
    - Kaqchikel, Akatenango Southwestern (`ckk`) [Mayan]. Kaqchikel (`cak`) already appears.

- Kaqchikel, Eastern (`cke`) [Mayan]. Kaqchikel (`cak`) already appears.
- Kaqchikel, Santa María de Jesús (`cki`) [Mayan]. Kaqchikel (`cak`) already appears.
- Kaqchikel, Southern (`ckf`) [Mayan]. Kaqchikel (`cak`) already appears.
- Kaqchikel, Western (`ckw`) [Mayan]. Kaqchikel (`cak`) already appears.
- Kaqchikel, Yepocapa Southwestern (`cbm`) [Mayan]. Kaqchikel (`cak`) already appears.
- K'iche', Joyabaj (`quj`) [Mayan]. K'iche' (`quc`) already appears.
- K'iche', West Central (`qut`) [Mayan]. K'iche' (`quc`) already appears.
- Mam, Southern (`mms`) [Mayan]. Mam (`mam`) already appears.
- Norwegian Bokmål (`nob`) [Indo-European]. Norwegian (`nor`) already appears.
- Norwegian Nynorsk (`nno`) [Indo-European]. Norwegian (`nor`) already appears.
- Nuclear Dan (`dnj`) [Mande]. Dan (`daf`) already appears.
- Poqomchi', Western (`pob`) [Mayan]. Poqomchi' (`poh`) already appears.
- South Transvaal Ndebele (`nbl`) [Atlantic-Congo]. Zulu (`zul`) already appears.
- Southern Rincon Zapotec (`zsr`) [Otomanguean]. Rincón Zapotec (`zar`) already appears.

### 1.1.2 Assigning linguistic families to languages in UD

Since this dataset is indexed by language names rather than standardized codes, additional care was required to resolve ambiguities, naming inconsistencies, and mismatches with existing classifications in Glottolog.

- **Languages where all frequencies are 0, removed.** After extracting word-order information from the UD datasets for the following languages, no S, O and V triplet could be obtained in the following samples:

  - Old Turkish [Turkic]

- **Macrolanguages without Glottolog entries, removed or classified under the families of their constituent languages.** These languages correspond to macrolanguages, but they do not have entries in Glottolog. If none of the languages they encompass appears separately in the data collection, they are classified under the highest-level language family. This is the case of

  - Azerbaijani [Turkic]

- Persian [Indo-European]
- Yiddish [Indo-European]

If any of their languages appears separately, they are removed. The removed macrolanguage samples and the reason of the removal are the following:

- Chinese [Sino-Tibetan]. Cantonese, Classical Chinese, and Shanghainese already appear.

- **Macrolanguages listed in Glottolog as low-level families, removed or classified under the highest-level family.** These languages correspond to macrolanguages with entries in Glottolog, where they are treated as low-level families rather than as individual languages. If none of the languages they cover appears separately in the data collection, they are classified under the highest-level language family. This is the case of

    - Pashto [Indo-European]
    - Uzbek [Turkic]

    If any of their languages appears separately, they are removed. The removed low-level family samples and the reason of the removal are the following:

    - Albanian [Indo-European]. Gheg already appears.
    - Arabic [Afro-Asiatic]. Maltese and South Levantine Arabic already appear.

- **Languages listed in Glottolog as dialects, removed or classified under the family of their language.** These languages appear in Glottolog, but their entries are treated as dialects of larger language units. If their language does not appear separately in the data collection, they are classified into the family of their language. These dialects and their solutions are the following ($\rightarrow$ indicates the language to which each dialect belongs):

    - Bokota → Buglere [Chibchan]
    - Cantonese → Yue Chinese [Sino-Tibetan]
    - Croatian → Serbian-Croatian-Bosnian [Indo-European]
    - Guajajara → Tenetehara [Tupian]
    - Khunsari → Khunsaric [Indo-European]
    - Luxembourgish → Moselle Franconian [Indo-European]
    - Nayini → Nayinic [Indo-European]
    - Neapolitan → Continental Southern Italian [Indo-European]
    - Serbian → Serbian-Croatian-Bosnian [Indo-European]

- Shanghainese → Wu Chinese [Sino-Tibetan]
- Soi → Soic [Indo-European]
- South Levantine Arabic → Levantine Arabic [Afro-Asiatic]
- Zaar → Saya [Afro-Asiatic]

If their language appears separately, they are removed. The removed dialect samples and the reason of the removal are the following:

- Latgalian [Indo-European]. Latvian already appears.
- Middle French [Indo-European]. French already appears.
- Ottoman Turkish [Turkic]. Turkish already appears.
- Pomak [Indo-European]. Bulgarian already appears.

- **Languages marked in UD as "code switching", removed.** The following entries correspond to systematic alternations between two distinct languages (e.g., Telugu and English) within a single discourse or text:

  - Frisian Dutch
  - Maghrebi Arabic French
  - Telugu English
  - Turkish English
  - Turkish German

  They were excluded from the analysis because they may not represent a single linguistic system with a stable word-order frequency. Furthermore, their inclusion would introduce a confounding factor, as the grammatical structures of the constituent languages are already represented in their respective individual entries.

- **Manually classified languages with ISO code, due to name mismatch with Glottolog.** Certain languages did not match any entry in Glottolog, either as a language or a dialect. However, the UD site provided a corresponding ISO 639-3 code. The following languages were manually classified under a standard Glottolog entry (→ points to the standard entry used):

  - Alemannic → Central Alemannic [Indo-European]
  - Ancient Greek → Ionic-Attic Ancient Greek [Indo-European]
  - Apurina → Apurinã [Arawakan]
  - Assyrian → Assyrian Neo-Aramaic [Afro-Asiatic]
  - Buryat → Russia Buriat [Mongolic-Khitan]
  - Cappadocian → Cappadocian Greek [Indo-European]
  - Classical Armenian → Classical-Middle Armenian [Indo-European]

- Egyptian → Egyptian (Ancient) [Afro-Asiatic]
- Gheg → Gheg Albanian [Indo-European]
- Gwichin → Gwich'in [Athabaskan-Eyak-Tlingit]
- Kaapor → Urubú-Kaapor [Tupian]
- Karo → Karo (Brazil) [Tupian]
- Khoekhoe → Nama (Namibia) [Khoe-Kwadi]
- Kiche → K'iche' [Mayan]
- Komi Permyak → Komi-Permyak [Uralic]
- Komi Zyrian → Komi-Zyrian [Uralic]
- Low Saxon → Eastern Low German [Indo-European]
- Makurap → Makuráp [Tupian]
- Mbya Guarani → Mbyá Guaraní [Tupian]
- Munduruku → Mundurukú [Tupian]
- Naija → Nigerian Pidgin [Indo-European]
- Nenets → Tundra Nenets [Uralic]
- Nheengatu → Nhengatu [Tupian]
- North Sami → North Saami [Uralic]
- Old East Slavic → Old Russian [Indo-European]
- Old English → Old English (ca. 450–1100) [Indo-European]
- Old French → Old French (842–ca. 1400) [Indo-European]
- Old Irish → Early Irish [Indo-European]
- Old Occitan → Old Provençal [Indo-European]
- Pesh → Pech [Chibchan]
- Skolt Sami → Skolt Saami [Uralic]
- Tupinamba → Tupinambá [Tupian]
- Western Sierra Puebla Nahuatl → Zacatlán-Ahuacatlán-Tepetzintla Nahuatl [Uto-Aztecan]
- Xavante → Xavánte [Nuclear-Macro-Je]
- Yakut → Sakha [Turkic]
- Yupik → Central Siberian Yupik [Eskimo-Aleut]

- **Manually classified languages without ISO code, due to name mismatch or ambiguous variants.** Some languages required manual classification because their names did not match any entry in Glottolog, or because they corresponded to geographical or dialectal variants for which no ISO code was specified. In cases of doubt, the descriptions provided on the UD site were used to identify the correct Glottolog entry. The following languages were reclassified (→ points to the language or standard entry used):

7

- Armenian → Eastern Armenian [Indo-European]
- Greek → Modern Greek [Indo-European]
- Guarani → Paraguayan Guaraní [Tupian]
- Haitian Creole → Haitian [Indo-European]
- Hebrew → Modern Hebrew [Afro-Asiatic]
- Indonesian → Standard Indonesian [Austronesian]
- Kyrgyz → Kirghiz [Turkic]
- Naga → Suansu [Sino-Tibetan]
- Old Church Slavonic → Church Slavic [Indo-European]
- Uyghur → Uighur [Turkic]

## 2 Methods

### 2.1 Overview

The philosophy of our methods is as follows. First, it is not a purely exploratory analysis. It is research driven by a theoretical prediction of swap distance minimization that has already been confirmed in different contexts [3–6]. Second, it addresses the problem of undersampling of word order frequencies by not targeting $\langle d \rangle$ but the difference between $\langle d \rangle$ and $\langle d \rangle_r$, where $\langle d \rangle_r$ preserves the distribution of word order probabilities. If certain word orders have zero probability because of undersampling, $\langle d \rangle_r$ takes it into account. Third, it is non-parametric. The main statistical test for $\langle d \rangle < \langle d \rangle_r$ within families (or languages lacking a dominant order) is non-parametric. The procedure to adjust the $p$-values of the statistical test is also non-parametric and has minimal assumptions [7]. Fourth, we avoid the choice of a significance level and claims about significance following recent recommendations [8]. The aim of our methodology is to show that we are able to get $p$-values smaller than recommended significance levels, e.g., 0.005 or 0.001 [9–11], when families are represented by a sufficiently large number of languages as in NT.

### 2.2 Alternatives to the main statistical test

The core statistical test for swap distance minimization is a one-tailed Wilcoxon signed-rank test. Notice that we could use other non-parametric tests, e.g., a permutation test on the difference between the mean $\langle d \rangle$ and the mean $\langle d \rangle_r$, but notice that such a kind of test would be based on a null hypothesis where all the values of $\langle d \rangle$ and $\langle d \rangle_r$ are merged losing the original matching between $\langle d \rangle$ and $\langle d \rangle_r$ in each language. The null hypothesis of the Wilcoxon signed-rank test preserves the matching, exchanging the values of $\langle d \rangle$ and $\langle d \rangle_r$ within each language. That is, the Wilcoxon signed-rank test performs a more conservative randomization of the data and is *a priori* less biased towards small $p$-values.

We do not test for swap distance minimization within individual languages for two reasons. First, it has been demonstrated mathematically that such a test would lack statistical power [5]. Second, we must control for vertical transmission (evolutionary relatedness) between languages. Linguistic families are seen as statistically independent if horizontal transmission (diffusion) is neglected [12]. Therefore, we run a statistical test for swap distance minimization within each family of an ensemble of linguistic families and verify *a posteriori* that the families with low *p*-values cover far regions of the world to ensure that horizontal transmission is less likely to explain our findings. The signed-rank test yields a *p*-value that indicates the likelihood that the difference between $\langle d \rangle$ and $\langle d \rangle_r$ for each language in a linguistic family is due to chance.

## 2.3 *p*-value adjustment

When we apply the main statistical test (Wilcoxon-signed rank test) to individual families, we apply a correction to the *p*-values to control for multiple comparisons. Probably, the two most popular methods are the Bonferroni correction and Holm-Bonferroni correction, which is a step-down procedure that is less conservative than the original Bonferroni correction. These methods assume that the distribution of the *p*-values being corrected is uniform in the interval $[0, 1]$ under the null hypothesis, which implies that the null hypothesis can always be rejected. However, under the null hypothesis of our Wilcoxon signed-rank test, the *p*-value follows some distribution on $[p_{min}, 1]$, where $p_{min}$ is a number that decreases as the number of languages in the family increases. If a family has a single language, $p_{min} \geq 0.5$. As the majority of our families are represented by a low number of languages, the application of Holm-Bonferroni turns out to be conservative because the *p*-values cannot be small in the many families with a few languages. Defining certain thresholds may help us to gain statistical power:

1. A threshold for the number of languages in a family to be included. That would reduce the penalty of the Holm-Bonferroni correction. In a traditional one-sided permutation test, the smallest *p*-value that can be achieved is $1/n!$ [13]. We are including many families that have just a few languages.

2. A threshold for the number of S, O and V triplets per language. For instance, [14] wrote: "we choose to eliminate corpora with fewer than 1,000 sentences, since we consider them too small to be representative of the language."

However, that approach implies defining arbitrary thresholds that become parameters of our *p*-value correction. Accordingly, we choose a parameter-free adjustment method that adapts to the distribution of *p*-values of each family under the null hypothesis of the Wilcoxon signed-rank test. This takes us to the minP family of correction methods [15]. A simple approach is the minP method, that is the analog of the Bonferroni correction. A less conservative

9

approach is the step-down minP (sd.minP) method, that is the analog of the well-known Holm-Bonferroni correction. A useful summary of these two methods can be found in Section "Resampling-Class Methods" of [16], where their outcomes correspond to our families and the two groups correspond to the original $\langle d \rangle$ and the random baseline ($\langle d \rangle_r$). These corrections require the generation of $N_{boot}$ observation vectors of size $M$, where $M$ is the number of families in the ensemble. The vector contains the $p$-values of the test for each family after randomizing the values of $\langle d \rangle$ and $\langle d \rangle_r$ for the languages within the family according to the null hypothesis of the Wilcoxon signed-rank test, i.e. that the values of $\langle d \rangle$ and $\langle d \rangle_r$ of a language are interchangeable. Thus, our randomization consists of swapping the values of $\langle d \rangle$ and $\langle d \rangle_r$ of each language with probability 1/2.

[15] recommends $N_{boot} = 10^4$. As this number may not be enough in practice, [16] used $N_{boot} = 10^5$. We used $N_{boot} = 10^6$.

Notice that our $p$-value adjustment procedure does not assume independence [15]. Although our test is applied on linguistic families and linguistic families are seen as statistically "independent", horizontal transmission (diffusion) may be a source of non-independence. Therefore, it is important that our adjustment method tolerates non-independence.

## 2.4 The problem of the estimation of $\langle d \rangle$

A problem in the estimation of $\langle d \rangle$ is the estimation of the $p_i$'s, namely the probabilities of each of the word orders, that are approximated by the relative frequency of each order in a sample. If the samples of a language are not large enough, certain estimated probabilities may be zero while they are not zero indeed [17, Table 1 and Table 2]. Furthermore, we cannot warrant that sampling effort is uniform across all languages. This is why we do not look for evidence of swap distance minimization directly on $\langle d \rangle$, but by comparing $\langle d \rangle$ against $\langle d \rangle_r$ for each language. $\langle d \rangle_r$ addresses the following question: what would be the expected value of $\langle d \rangle$ *in the same sample* if there was no swap distance effect? Therefore our method is robust to undersampling within languages because $\langle d \rangle_r$ takes into account critical characteristics of the sample such as its size or the number of non-zero probability orders.

 
}{}

\end{document}